\documentclass[sn-mathphys,Numbered]{sn-jnl}

\usepackage{graphicx}%
\usepackage{multirow}%
\usepackage{amsmath,amssymb,amsfonts}%
\usepackage{amsthm}%
\usepackage{mathrsfs}%
\usepackage[title]{appendix}%
\usepackage{xcolor}%
\usepackage{textcomp}%
\usepackage{manyfoot}%
\usepackage{booktabs}%
\usepackage{algorithm}%
\usepackage{algorithmicx}%
\usepackage{algpseudocode}%
\usepackage{listings}%
\usepackage{graphicx}
\usepackage{array}
\usepackage{multirow}
\usepackage[pagewise]{lineno}
\usepackage{booktabs}
\usepackage[normalem]{ulem}
\usepackage{natbib}
\usepackage{caption}
\usepackage{caption}
\usepackage{amsmath}
\usepackage{framed}
\usepackage{mdframed}
\usepackage{lipsum}
\usepackage{hyperref}
\usepackage{subcaption}
\usepackage{footnote}
\usepackage{enumitem}
\usepackage{longtable}
\usepackage{color}
\usepackage{adjustbox}
\usepackage{marvosym}
\colorlet{punct}{red!60!black}
\definecolor{background}{HTML}{EEEEEE}
\definecolor{delim}{RGB}{20,105,176}
\colorlet{numb}{magenta!60!black}

\usepackage{blindtext}
\usepackage{xparse}
\DeclareDocumentCommand\dia{ o m }{%
    \begin{itemize}[%
        ,label=\IfNoValueTF {#1} {}{#1:}
        ,labelsep=8mm
        ,nosep 
        ,font=\color{black}
        ]
        \item #2
    \end{itemize}%
    }

\theoremstyle{thmstyleone}%
\theoremstyle{thmstyletwo}%

\theoremstyle{thmstylethree}%
\newcolumntype{M}[1]{>{\centering\arraybackslash}m{#1}}

\definecolor{effort}{HTML}{6DFCFF}
\definecolor{outcome}{HTML}{F4CCCC}
\definecolor{royal_purple}{RGB}{153, 21, 78}
\DeclareCaptionLabelSeparator{custom}{. }
\begin{document}

\title[Article Title]{How Can I Get It Right?  Using GPT to Rephrase Incorrect Trainee Responses}


\author*[1]{\fnm{Jionghao} \sur{Lin}}\email{jionghao@cmu.edu}

\author[1]{\fnm{Zifei} \sur{Han}}\email{hanzifeifei@gmail.com }

\author[1]{\fnm{Danielle R.} \sur{Thomas}}\email{drthomas@cmu.edu}

\author[1]{\fnm{Ashish} \sur{Gurung}}\email{agurung@andrew.cmu.edu }

\author[1]{\fnm{Shivang} \sur{Gupta}}\email{shivang@cmu.edu}

\author[1]{\fnm{Vincent} \sur{Aleven}}\email{aleven@cs.cmu.edu}

\author[1]{\fnm{Kenneth R.} \sur{Koedinger}}\email{koedinger@cmu.edu }

\affil[1]{\orgdiv{Human-Computer Interaction Institute}, \orgname{Carnegie Mellon University}, \orgaddress{\street{5000 Forbes Ave}, \city{Pittsburgh}, \postcode{15213}, \state{PA}, \country{USA}}}











\abstract{One-on-one tutoring is widely acknowledged as an effective instructional method, conditioned on qualified tutors. However, the high demand for qualified tutors remains a challenge, often necessitating the training of novice tutors (i.e., trainees) to ensure effective tutoring. Research suggests that providing timely explanatory feedback can facilitate the training process for trainees. However, it presents challenges due to the time-consuming nature of assessing trainee performance by human experts. Inspired by the recent advancements of large language models (LLMs), our study employed the GPT-4 model to build an explanatory feedback system. This system identifies trainees' responses in binary form (i.e., correct/incorrect) and automatically provides template-based feedback with responses appropriately rephrased by the GPT-4 model. We conducted our study on 410 responses from trainees across three training lessons: \textit{Giving Effective Praise}, \textit{Reacting to Errors}, and \textit{Determining What Students Know}. Our findings indicate that: 1) using a few-shot approach, the GPT-4 model effectively identifies correct/incorrect trainees' responses from three training lessons with an average F1 score of 0.84 and an AUC score of 0.85; and 2) using the few-shot approach, the GPT-4 model adeptly rephrases incorrect trainees' responses into desired responses, achieving performance comparable to that of human experts.}




\maketitle

\section{Introduction}\label{sec1}

One-on-one tutoring has been recognized as a highly effective strategy for enhancing student learning, with substantial evidence supporting its impact~\cite{kraft2021blueprint, nickow2020impressive}. However, there are significant challenges associated with the scalability of one-on-one tutoring, primarily due to the scarcity of skilled tutors, including certified teachers and paraprofessionals. This shortage has left an estimated 16 million students in the United States in need of individualized support, as highlighted by~\cite{kraft2021blueprint}. In response to this shortage, there has been a strategic shift towards effectively training novice tutors, including community volunteers, retired individuals, and college students, to fulfill tutoring role~\cite{nickow2020impressive}.

\begin{figure*}[b]
\centering
  \includegraphics[width=0.99\textwidth]{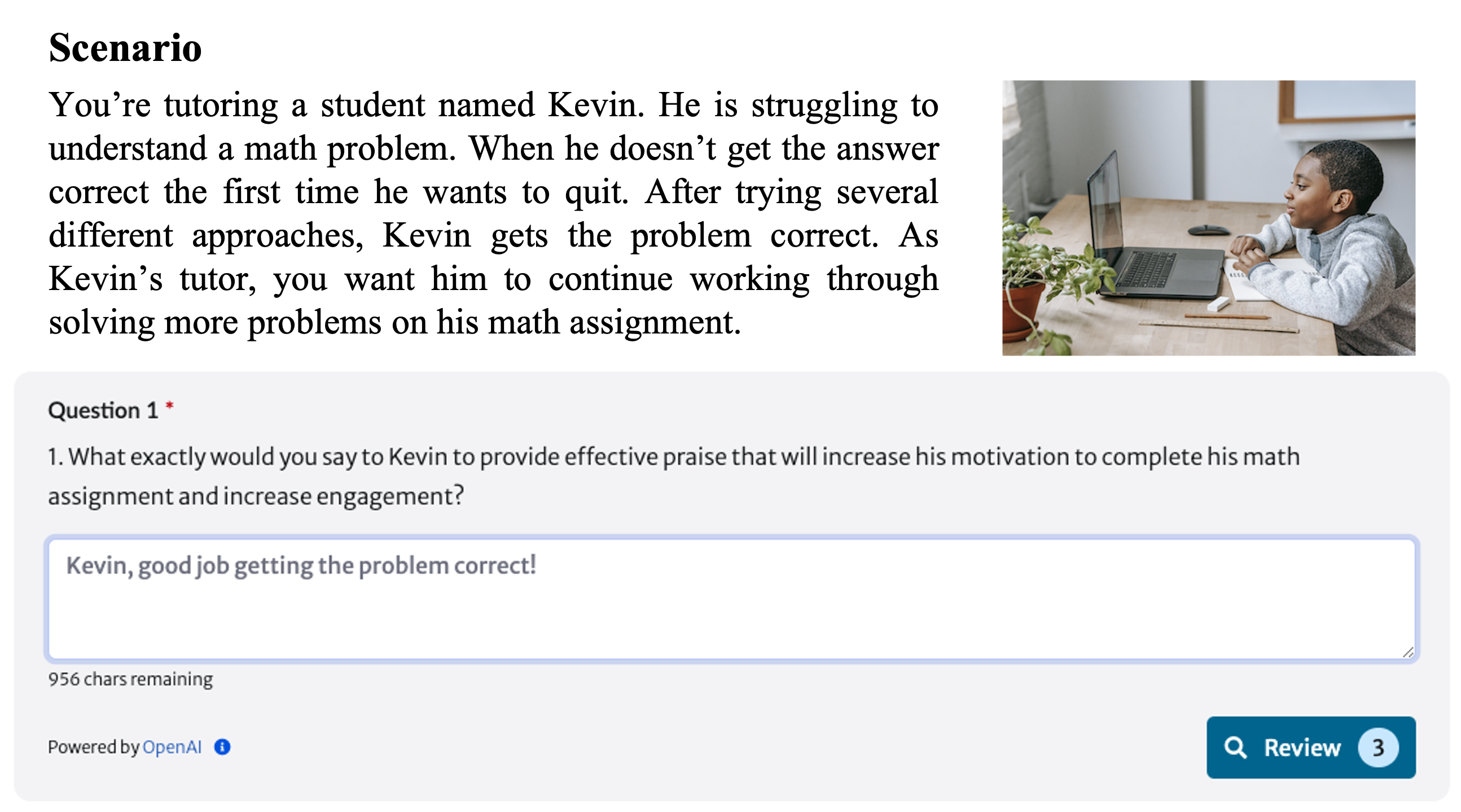}
\caption{ An example of a trainee (i.e., novice tutor) incorrectly responding to an open-ended question on how to best reply to a student by giving effective praise. In this particular example, the trainee is praising the student for getting the problem correct, which is achievement or outcomes-based praise and not based on effort.}
\vspace{-5mm}
\label{fig:intro_scenario}
\end{figure*}

The growing demand for skilled tutors has resulted in the development of various professional development programs tailored to the unique needs of nonprofessional and novice tutors~\cite{nickow2020impressive}. Driven by this need, researchers have explored the use of online scenario-based training to simulate real-life tutoring scenarios for novice tutors~\cite{thomas2023tutor} and pre-service teachers~\cite{thompson2019teacher}. Fig. \ref{fig:intro_scenario} illustrates a scenario on \textit{Giving Effective Praise}. It demonstrates how tutors can fail to appropriately acknowledge the student's efforts by providing outcome-based praise as opposed to effort-based praise. For instance, saying \textit{``Kevin, good job getting the problem correct!''} fails to acknowledge the student's efforts and persistence. As indicated in previous research \cite{lin2023using, hirunyasiri2023comparative}, the availability of real-time explanatory feedback within the scenario-based training lessons can help tutors provide effective praise. Particularly, real-time feedback on learners' errors, similar to the feedback received while engaging in the deliberate practice of responding to situational judgment tests, is described as a favorable learning condition and can lead to better learning outcomes~\cite[p.~5]{koedinger2023astonishing}.

While the benefits of real-time explanatory feedback in enhancing tutor learning outcomes are well-documented, crafting such feedback presents substantial challenges due to its labor-intensive nature. Traditionally, providing this level of specialized training, replete with personalized explanatory feedback, warrants a substantial investment of effort and time. The process of providing personalized feedback to novice tutors requires considerable time and effort from skilled tutors to ensure feedback effectiveness and relevance. Moreover, beyond the substantial investment of time and effort, the feasibility of scaling such training protocols to meet the high demand across educational settings significantly compounds the challenge. However, recent breakthroughs in large language models (LLMs) offer a promising avenue for streamlining this process. Models such as the Generative Pre-trained Transformer (GPT) could potentially automate the generation of personalized, real-time feedback for tutors~\cite{hirunyasiri2023comparative, dai2023can}. This automation not only has the potential to alleviate the resource burden but also to enhance the specificity and precision of the feedback by accurately identifying the personalized needs of the tutors~\cite{hirunyasiri2023comparative}.



Currently, the quality of automated explanatory feedback is lacking, with many systems failing to provide learners with accurate feedback on their constructed responses \cite{lin2023using, hirunyasiri2023comparative}.  We argue that the quality of feedback for tutor training can be further improved. Inspired by the feedback research~\cite{henderson2019impact, hattie2007power, butler2013explanation}, where learners interpret performance-related information to enhance their understanding, we postulate that presenting desired tutoring responses within feedback to novice tutors can enhance the effectiveness of the training. However, rephrasing incorrect tutor responses into the correct or desired form often necessitates a substantial investment of time and effort from experienced tutors—hence introducing scalability constraints associated with tutor training. Thus, we aim to explore approaches to improve our ability and accuracy in providing tutors with explanatory feedback while also mitigating the time and effort requirements of human graders by automating the process of generating explanatory feedback and correction to their responses. The automation requires the development of classification systems that can effectively analyze tutor responses or, in other words, classification systems that determine the correctness of tutor responses to scenario-specific requirements of the learners. However, there is useful learner information within appropriate classified incorrect responses. These incorrect learner-sourced responses can be used to provide tutors corrective, explanatory feedback by taking an incorrect response and rephrasing or modifying it to make it a desired, or correct, response. Research supports  when learners are given specific feedback related to their responses, such as taking incorrect tutor responses and personalizing them by making them correct, they gain a better understanding of their learning \cite{attali2010immediate, torres2022feedback}.  

We aim to explore how GPT models can serve as supplementary tools to deliver synchronous feedback to tutors on their responses of how to best respond to specific training scenarios (e.g., praising a student for effort) leveraging useful tutor incorrect responses. We propose two \textbf{R}esearch \textbf{Q}uestions:

\vspace{1mm}

\begin{itemize}[leftmargin=.48in]
    \item[\textbf{RQ1: }] Can a large language model accurately identify trainees' incorrect responses where trainees failed to effectively guide students in specific training scenarios?

    \item[\textbf{RQ2: }] Can GPT-4 be harnessed to enhance the effectiveness of trainees' responses in specific training scenarios?
\end{itemize}

\vspace{1mm}

\begin{figure*}[h]
\centering
  \includegraphics[width=0.85\textwidth]{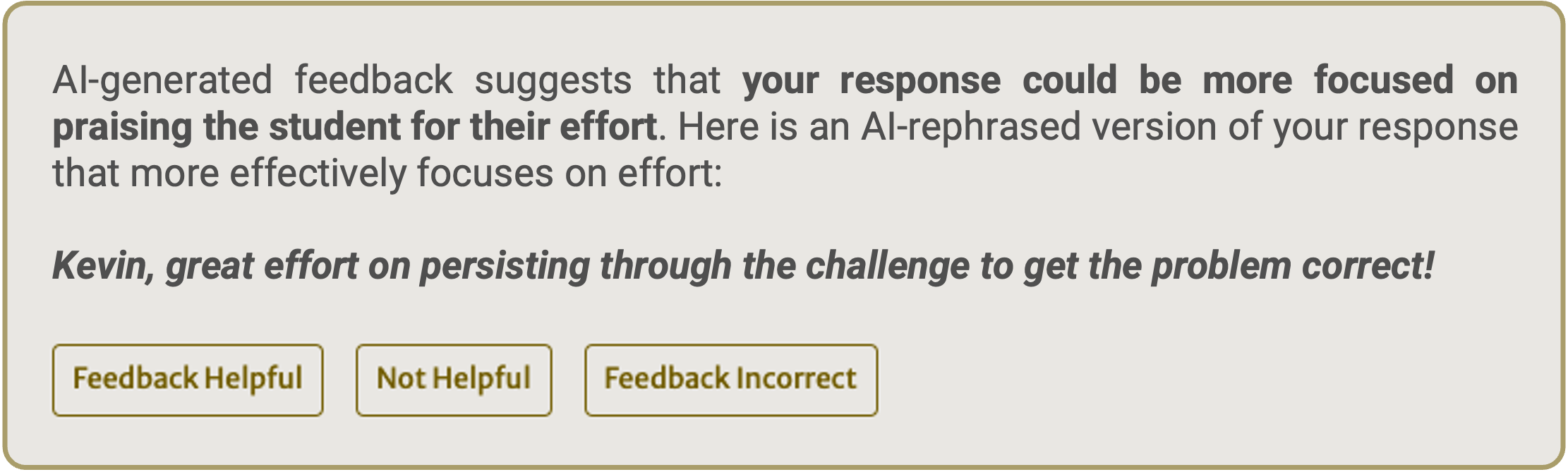}
\caption{Explanatory feedback for novice tutor responses.}
\label{fig:intro_tem_feedback}
\vspace{-5mm}
\end{figure*}

We initially developed a binary classifier to determine tutor's correct and incorrect responses from three training lessons: \textit{Giving Effective Praise}, \textit{Reacting to Errors}, and \textit{Determining What Students Know}. We employed zero-shot and few-shot learning approaches to classify the trainees' responses. Our result demonstrated that the five-shot learning approach achieved acceptable performance in identifying the incorrect responses. Building upon the results of \textbf{RQ1}, we selected the incorrect responses identified by our optimal few-shot learning classifier, which is further used for the \textbf{RQ2}. We explored the idea of rephrasing incorrect trainees' responses to determine if we can prompt GPT-4 to effectively make them correct. An example of an incorrect response from the lesson \textit{Giving Effective Praise} is shown in Fig. \ref{fig:intro_scenario}), e.g., \textit{``Kevin, good job getting the problem correct!''}. Through extensive experiments, we obtained an effective prompt to secure the rephrased responses presented in an accurate form with minimal changes of the words from the original incorrect responses. Building upon the result from RQ1 and RQ2, we build a feedback system to provide explanatory feedback to the incorrect trainee's response shown in Fig. \ref{fig:intro_tem_feedback}. 

\section{Related Work}\label{rw}

\subsection{Significance of Feedback on Learning}
Feedback plays a crucial role in improving the students' learning outcomes and performance~\cite{hattie2007power, henderson2019impact, lin2023learner}. In the field of feedback research, theoretical models have been developed to explain the impact of feedback on learning and to identify the core principles that underpin effective feedback design. \citeauthor{hattie2007power}~\cite{hattie2007power} defined feedback as the information about the correctness of a learner's actions or decisions, along with explanations about why those actions or decisions are right or wrong, underlines the significance of feedback. As emphasized in their work \cite{hattie2007power}, the influence of feedback on learning varies based on the type and timing of its delivery. 

Effective feedback should assist learners in understanding the rationale behind the feedback, which is crucial for deeper learning \cite{henderson2019impact}. Moreover, including the correct answer within the feedback substantially enhances its efficacy by offering learners the information needed to correct their errors \cite{butler2013explanation}. This is especially relevant when learners answer open-ended questions, as simply knowing that their response is incorrect may not suffice to improve their understanding \cite{butler2013explanation}. By presenting the correct answer (or correct responses to open-ended question) in the feedback, learners can compare their responses with the correct responses, identify areas for improvement, and gain guidance on how to approach similar questions in the future \cite{attali2010immediate, torres2022feedback}. To help learners identify their misconception in the open-ended question, we posit that it is necessary to include the correct responses in the feedback. However, providing timely explanatory feedback faces challenges since crafting effective explanatory feedback is often time-consuming and labor-intensive nature \cite{lin2023learner, dai2023can, hirunyasiri2023comparative}. To address this issue, it is necessary to develop automated feedback generation system.

\subsection{Feedback Generation}
The development of automated feedback has received significant attention from educational researchers \cite{dai2023can, lin2023using, hirunyasiri2023comparative, pardo2018ontask, demszky2021can}. For example, Ontask~\cite{pardo2018ontask} is a rule-based feedback provision system designed to assist instructors in delivering personalized feedback based on specific conditions of learners (e.g., the duration spent on the learning system). Additionally, \citeauthor{demszky2021can}~\cite{demszky2021can} developed a feedback system that automatically delivers explanatory feedback to instructors via email within two to four days after their tutoring sessions. Their study results \cite{demszky2021can} indicate that timely explanatory feedback enhanced learners' satisfaction. \citeauthor{lin2023using}~\cite{lin2023using} used sequence labeling techniques to provide automated explanatory feedback, which demonstrated the potential of the large language models on identifying the effective components of feedback. Despite demonstrating the effectiveness of automated feedback systems, the provision of feedback with correct responses to open-ended question is still under-explored, which are needed to advance feedback systems

\subsection{Using Large Language Models for Feedback Generation}

Inspired by recent research on using large language models for feedback generation~\cite{levonian2023retrieval, lin2023using, hirunyasiri2023comparative, mcnichols2023exploring, macneil2022generating, daiaassessing}, we posit that GPT-based large language models hold potential for advancing the development of automated feedback. For example, \citeauthor{dai2023can} \cite{dai2023can} investigated the capability of GPT-3.5 model (ChatGPT) to generate feedback for students' writing assignment and they \cite{dai2023can} found that GPT-3.5 could produce feedback that was more readable than that of human instructors. Subsequently, \citeauthor{daiaassessing} \cite{daiaassessing} found that GPT-4 outperformed both GPT-3.5 and human instructors in providing effective feedback based on the feedback attributes proposed by \cite{hattie2007power}. Then, \citeauthor{hirunyasiri2023comparative}~\cite{hirunyasiri2023comparative} leveraged the GPT-4 model to provide timely feedback for human tutors' training. Their results \cite{hirunyasiri2023comparative} indicated that GPT-4 outperformed human educational experts in identifying a specific tutoring practice, giving effective praise. While these studies have demonstrated the feasibility of GPT-based models in feedback generation, none have ventured into generating explanatory feedback with correct responses to open-ended questions. Given that GPT-4 has shown remarkable performance on various educational tasks (e.g., generating high-quality answer responses for middle school math questions \cite{levonian2023retrieval} and providing feedback for multiple-choice questions at the middle-school math level~\cite{mcnichols2023exploring}), our study also leveraged the GPT-4 model to further explore its capabilities in automatically generating explanatory feedback.



\section{Method}\label{method}

\subsection{Data}
\label{sec:data}
We developed an online learning platform\footnote{\url{https://www.tutors.plus/solution/training}} to facilitate training for the novice tutors in the form of brief scenario-based lessons. Within the scope of this study, we refer to the novice tutors participating in the training activities as \textit{trainees}. Aligning with previously demonstrated competencies of effective tutoring \cite{chhabra2022evaluation}, each lesson presents scenario-based questions to facilitate an authentic and contextually relevant tutor learning opportunity. These scenarios challenged the tutors to apply their knowledge and skills by simulating real-world tutoring situations (see Fig.~\ref{fig:intro_scenario}).  We examined the trainees' performance and understanding across three lessons: \textit{Giving Effective Praise}, \textit{Reacting to Errors}, and \textit{Determining What Students Know}. These lessons are based on the skillsets that were identified to be crucial for tutors in previous work \cite{chhabra2022evaluation, thomas2023tutor}.



Each lesson consisted of two scenarios. Across all trainees, we collected 410 responses: 140 responses from the 70 trainees who took the \textit{Giving Effective Praise} lesson, 118 responses from \textit{Reacting to Errors} (59 trainees), and 152 responses from \textit{Determining What Students Know} (76 trainees). Before analysis, we removed 10, 4, and 13 responses respectively from each lesson because they were either empty or contained incoherent or meaningless content (e.g., ``\textit{ad;fajkl}'', ``\textit{test test test}'' or ``\textit{I have no idea}''), resulting in a total of 383 analyzed responses. We also collected demographic information about the trainees, including their experience as tutors, as presented in Table \ref{tab:demo}. For each lesson, tutors provided self-reported demographic details, including information regarding their race, gender, age, and tutoring experience.

\begin{table}[!ht]
\footnotesize
\caption{Demographic information of participants}
\label{tab:demo}
\renewcommand{\arraystretch}{1.3}
\begin{tabular}{lccc}
\hline
\textbf{\begin{tabular}[c]{@{}l@{}}Demographic \\ Categories\end{tabular}} &
  \textit{\begin{tabular}[c]{@{}c@{}}Giving Effective \\ Praise (n = 70)\end{tabular}} &
  \textit{\begin{tabular}[c]{@{}c@{}}Reacting to \\ Errors (n = 59)\end{tabular}} &
  \textit{\begin{tabular}[c]{@{}c@{}}Determining What \\ Students Know (n = 76)\end{tabular}} \rule{0pt}{4.6ex}\rule[-4.2ex]{0pt}{0pt} \\ \hline
\textbf{Gender}              &    &    &      \\
\hspace{1em}Female                       & 27 & 23 & 32 \\
\hspace{1em}Male                         & 34 & 30 & 34 \\
\hspace{1em}Non-binary                   & 1  & 1  & 1   \\
\hspace{1em}Unknown                      & 8  & 5  & 9   \\
\textbf{Age}                 &    &    &      \\
\hspace{1em}18-24                        & 8  & 8  & 11  \\
\hspace{1em}25-34                        & 11 & 6  & 9  \\
\hspace{1em}35-50                        & 12 & 9  & 9  \\
\hspace{1em}51-64                        & 21 & 22 & 25 \\
\hspace{1em}65+                          & 12 & 12 & 16 \\
\hspace{1em}Unknown                      & 6  & 2  & 6 \\
\textbf{Ethnicity}           &    &    &       \\
\hspace{1em}Asian                        & 12 & 10 & 18 \\
\hspace{1em}White         & 34 & 30 & 34  \\
\hspace{1em}Others                       & 8  & 8  & 9  \\
\hspace{1em}Unknown                      & 16 & 11 & 15 \\
\textbf{\begin{tabular}[c]{@{}l@{}}Tutoring \\ Experience*\end{tabular}} &    &    &       \\
\hspace{1em}Level 1                      & 4  & 3  & 4  \\
\hspace{1em}Level 2                      & 14 & 11 & 14 \\
\hspace{1em}Level 3                      & 22 & 18 & 24 \\
\hspace{1em}Level 4                      & 20 & 21 & 24 \\
\hspace{1em}Level 5                      & 5  & 4  & 5 \\
\hspace{1em}Unknown                      & 5  & 6  & 5 \\ 
\hline
\end{tabular}
\textbf{Tutoring Experience*}: The tutors were asked to rate their prior tutoring experience on a five-point Likert scale, where Level 1 indicated a tutor with limited experience, and Level 5 signified an expert tutor.
\end{table}


\subsection{Annotation for Trainee's Responses}
\label{annota_trainee}
In the lesson \textit{Giving Effective Praise}, trainees practice their skills in engaging students by offering effort-based praise. The praise provided by trainees should effectively acknowledge students' efforts and aim to enhance their motivation and desire to keep learning. A tutoring scenario was depicted where a student was struggling to persevere on an assignment (See the scenario in Table \ref{tab:sample_praise}). The tutor trainee's responses were expected to show the components of effective praise as suggested by research recommendation~\cite{thomas2023tutor}. Effective praise should be: 1) timely, positive, and sincere, 2) highlighting what student did well during the tutoring, 3) genuine and avoiding generic comments like \textit{``great job''}, and 4) focus on the learning process rather than on the student or the outcome. In short, correct praise responses should be supportive, positive, encouraging, and acknowledging the student's effort during the learning process. In Table \ref{tab:sample_praise}, we demonstrate some praise responses with an explanation of the rationale for labeling responses as either \textbf{Correct} or \textbf{Incorrect}.

\begin{table}[h]
\caption{Examples of correct and incorrect trainee responses for the lesson \textit{Giving Effective Praise} with annotation rationale.}
\label{tab:sample_praise}
\footnotesize
\renewcommand{\arraystretch}{1.35}
\begin{tabular}{|ll|}
\hline
\multicolumn{2}{|l|}{\textbf{Scenario}} \rule{0pt}{3.6ex}\rule[-2.2ex]{0pt}{0pt}\\ \hline
\multicolumn{2}{|l|}{\begin{tabular}[c]{@{}l@{}} \textit{You're tutoring a student named Kevin. He is struggling to understand a math problem. When} \\ \textit{he doesn't get the answer correct the first time, he wants to quit. After trying several different} \\ \textit{approaches, Kevin gets the problem correct. As Kevin's tutor, you want him to continue working} \\ \textit{through solving more problems on his math assignment.}\end{tabular}} \\ \hline
\textbf{Trainee Response}         & \textbf{Interpretation}       \rule{0pt}{3.6ex}\rule[-2.2ex]{0pt}{0pt} \\ \hline
\textit{\begin{tabular}[c]{@{}l@{}}You are making steady progress and \\ it is good to see the results of your efforts\end{tabular}} &
  \begin{tabular}[c]{@{}l@{}}\textbf{Correct Response}\\ The response expresses the sense of positive and \\ sincere. The praise focuses on the student's \\ perseverance and acknowledges the students \\ for working hard and the process of learning.\end{tabular} \\ \hline
\textit{You did a great job, well done!} &
  \begin{tabular}[c]{@{}l@{}}\textbf{Incorrect Response}\\ This response is sincere and positive but the praise\\ does not focus on student learning efforts or\\  learning actions (e.g., demonstrated the \\ problem-solving procedural).\end{tabular} \\ \hline
\end{tabular}
\end{table}

In the lesson \textit{Reacting to Errors}, trainees practice their skills in responding to student errors. Trainees employ various pedagogical strategies aimed towards addressing gaps in the learners' knowledge through constructive feedback. Instead of overt criticism, the emphasis is on fostering a positive approach to errors. This approach seeks to shift students' perception towards errors by underscoring their importance in the learning process. A tutoring scenario was depicted where a student made a mistake in solving a problem (See the scenario in Table \ref{tab:sample_error}). The tutor trainee's responses to students' errors should help students develop their critical thinking skills and encourage students to correct their mistakes. According to~\cite{thomas2023tutor}, to effectively respond to students' errors, one should: 1) indirectly inform students about their mistake in the problem-solving process, 2) guide the student towards self-correction, and 3) show praise for the student's effort or attempt. Responses that directly highlight the student's error or inform the student what to do are not desired in the tutoring practice~\cite{thomas2023tutor}. In Table \ref{tab:sample_error}, we demonstrated some responses of reacting to errors with the explanation of the rationale for labeling responses as either \textbf{Correct} or \textbf{Incorrect}.

\begin{table}[h]
\caption{Examples of both correct and incorrect trainee responses for the lesson \textit{Reacting to Errors} with annotation rationale.}
\label{tab:sample_error}
\footnotesize
\renewcommand{\arraystretch}{1.35}
\begin{tabular}{|ll|}
\hline
\multicolumn{2}{|l|}{\textbf{Scenario}} \rule{0pt}{3.6ex}\rule[-2.2ex]{0pt}{0pt}\\ \hline
\multicolumn{2}{|l|}{\begin{tabular}[c]{@{}l@{}}\textit{Imagine you are a mentor to a student, Aaron, who has a long history of struggling with math.} \\ \textit{Aaron is not particularly motivated to learn math. He just finished a math problem adding a} \\ \textit{3-digit and 2-digit number and has made a common mistake (shown below).} \end{tabular}} \\ \hline
\textbf{Trainee Response:} &
  \textbf{Interpretation:} \rule{0pt}{3.6ex}\rule[-2.2ex]{0pt}{0pt}\\ \hline
\textit{\begin{tabular}[c]{@{}l@{}}Lucy, very well, but I have to point that\\  we have another way of doing the \\ math problem, we can repeat the math \\ together, what do you think?\end{tabular}} &
  \begin{tabular}[c]{@{}l@{}}\textbf{Correct Response}\\ This response avoids using direct words about \\ the student’s mistake and implicitly clears up \\ the misconception. Instead, the response \\ encourages the student to make another attempt \\ instead of explicit answers\end{tabular} \\ \hline
\textit{\begin{tabular}[c]{@{}l@{}}This is very close! I see one issue, \\ can you walk me through the how \\ you worked through the problem?\end{tabular}} &
  \begin{tabular}[c]{@{}l@{}}\textbf{Incorrect Response}\\ This response asks the student to walk through \\ the steps but it still uses the word “issue”,\\  which may be frustrating.\end{tabular} \\ \hline
\end{tabular}
\end{table}

In the lesson \textit{Determining What Students Know}, this lesson is designed to enhance the tutor trainees' skills in discerning the current knowledge level of the students by distinguishing what the students have comprehended and what still needs to be learned. A tutoring scenario was depicted where a student was given a math problem they did not know how to solve (see the scenario in Table \ref{tab:sample_know}). The tutor trainee's responses were used to gauge the student's prior knowledge at the start of the session and provide instruction based on what students already know as a launching point for the rest of the session. According to~\cite{thomas2023tutor}, effective response of determining what students know should be: 1) prompting students to demonstrate what they have already done or explain what they know, 2) presenting in an open-ended form and avoiding asking student's understanding of specific knowledge concept, 3) guiding the tutoring conversation to locate the student's misunderstanding, 4) providing instructional support to help students find the correct answer. To summarize, correct response of determining what students know should assess a student’s prior knowledge, guide the conversation to catch student's misconceptions or errors and support productive struggle. In Table \ref{tab:sample_know}, we demonstrated some responses of determining what students know with the explanation of the rationale for labeling responses as either \textbf{Correct} or \textbf{Incorrect}.

\begin{table}[h]
\caption{Examples of both correct and incorrect trainee responses for the lesson \textit{Determining What Students Know} with annotation rationale.}
\label{tab:sample_know}
\footnotesize
\renewcommand{\arraystretch}{1.35}
\begin{tabular}{|ll|}
\hline
\multicolumn{2}{|l|}{\textbf{Scenario}} \rule{0pt}{3.6ex}\rule[-2.2ex]{0pt}{0pt}\\ \hline
\multicolumn{2}{|l|}{\begin{tabular}[c]{@{}l@{}}\textit{You are working with a student named Cindy on her math homework. She is having trouble solving} \\ \textit{a geometry problem dealing with triangles. She shows you the following diagram displaying a triangle} \\ \textit{and states that she has to determine the value of angle x. Cindy says, "I don't know what to do."}\end{tabular}} \\ \hline
\textbf{Trainee Response:} &
  \textbf{Interpretation:} \rule{0pt}{3.6ex}\rule[-2.2ex]{0pt}{0pt}\\ \hline
\textit{What have you tried so far?} &
  \begin{tabular}[c]{@{}l@{}}\textbf{Correct Response}\\ This response asks an open-ended question \\ to understand what students have tried so far \\ and gauge the student's knowledge\end{tabular} \\ \hline
\textit{Do you know what PEMDAS means?} &
  \begin{tabular}[c]{@{}l@{}}\textbf{Incorrect Response}\\ Tutors’ responses can contain questions to \\ students but they must be open-ended and \\ non-specific to assess student’s knowledge \\ of an individual knowledge component.\end{tabular} \\ \hline
\end{tabular}
\end{table}

\subsection{Identifying desired trainee responses}
\label{method:rq1}




One of the motivations for this study is the creation of a classifier capable of discerning desired attributes in a tutor's responses to scenario-based prompts. The goal is to determine whether the tutors can adapt to the specific scenarios and integrate scenario-specific instructional practices when supporting the learners. For instance, should a trainee fail to acknowledge the learner's effort when working on an activity requiring effective praise, the classifier would categorize the tutor's feedback as \textbf{Incorrect} (less desirable). Identifying these scenarios presents an opportunity to personalize training activities for trainees, enhancing their ability to learn from and rectify specific instructional methodologies. 

In addressing \textbf{RQ1}, we first employed two expert raters, both specialists in educational instruction and feedback, to annotate trainees' responses as either \textbf{Correct} (desirable) or \textbf{Incorrect} (less-desirable). Using Cohen's $\kappa$, we determined inter-rater reliability, obtaining scores of 0.85, 0.81, and 0.64 for \textit{Giving Effective Praise}, \textit{Reacting to Errors}, and \textit{Determining What Students Know}, respectively. These scores of inter-rater reliability are considered sufficient \cite{neuendorf2017content}. Disagreements between the raters prompted input from a third expert to ensure consistency in annotations. Then, recognizing the typical need for a large amount of data when training classifiers from scratch for natural language processing tasks, we turned to recent advances in machine learning. As documented in~\cite{wang2020generalizing, pourpanah2022review}, zero-shot and few-shot learning methods can effectively discern patterns in datasets, even when they are limited or absent. These methods leverage the inherent capability of pre-trained models, which is crucial for ensuring classification performance and generalizability. The principle mirrors human cognition, as explored in~\cite{wang2020generalizing, pourpanah2022review}, where individuals apply their generalized knowledge to identify unfamiliar objects or concepts. Further details of these methods are described below:

\begin{itemize}
    \item \textbf{Zero-shot Learning:} In zero-shot learning, the classifier is trained to perform tasks for which it has seen no labeled examples at all. This is achieved by transferring knowledge from related tasks and using semantic relationships between classes. The model's prior knowledge, often in the form of embeddings or representations that capture semantic meanings, is crucial for it to make predictions in unseen classes~\cite{pourpanah2022review}.
    
    \item \textbf{Few-shot Learning:} In few-shot learning, the classifier is trained to perform tasks using a limited amount of labeled data. The underlying principle is to leverage the knowledge acquired by the model from previous and related tasks to facilitate effective generalization to a new task, even when provided with minimal data. The prior knowledge enables the classifier to adapt to new tasks with only a few examples~\cite{wang2020generalizing}. Additionally, given that our classifier is designed to categorize trainees' responses into two categories (i.e., correct or incorrect), the few-shot learning with two classification categories is commonly termed \textit{``two-way few-shot learning''}. For instance, a two-way two-shot contains two correct responses and two incorrect responses. Upon a thorough review of existing literature~\cite{cao2019theoretical}, we found that most studies implemented few-shot learning with the number of shots less than or equal to five. In line with this consensus, our study also sets five shots as the maximum threshold for the number of shots.
\end{itemize}

As described, both zero-shot and few-shot learning methods rely on a robust pre-trained model. These pre-trained models, having been exposed to extensive training corpora, inherently possess base knowledge that allows them to discern generalized patterns even from minimal datasets. Inspired by the effectiveness of GPT-4 models in existing educational tasks \cite{levonian2023retrieval, mcnichols2023exploring, hirunyasiri2023comparative}, we adopted the state-of-the-art GPT-4 model~\cite{openai2023gpt4} as the foundational model for conducting binary classification of trainees' responses. A GPT prompt is a sentence or phrase provided to the GPT model to produce a response~\cite{dai2023can, li2023can}. Our prompt strategies are detailed in Table \ref{tab:promt_bin}. 


\begin{table}[!ht]
\caption{Prompt strategies for a binary classifier. We used \textit{Chat-Completion} to process the trainees' responses in batch}
\footnotesize
\label{tab:promt_bin}
\renewcommand{\arraystretch}{1.5}
\begin{tabular}{|lp{4.2cm}|lp{4.2cm}|}
\hline
\multicolumn{2}{|l|}{\textbf{Zero Shot}}                   & \multicolumn{2}{l|}{\textbf{Few-shot}}                                                          \\ \hline
\textbf{Role}      & \textbf{Content}                              & \textbf{Role}      & \textbf{Content}                                               \\ \hline
\textbf{System}    & \textit{``You are a binary classifier.''}                  & \textbf{System}    & \textit{``You are a binary classifier.''}                                   \\
\textbf{User} &
  \texttt{\{Lesson Principle\}} + \textit{``According to the lesson principle, please determine if the following response contains''} + \texttt{\{Lesson Name\}} + \textit{``please respond YES; if not, please respond NO.''} &
  \textbf{User} &
  \texttt{\{Lesson Principle\}} + \textit{``According to the lesson principle, please determine if the following response contains''} + \texttt{\{Lesson Name\}} + \textit{``please respond YES; if not, please respond NO.''} \\
\textbf{Assistant} & \textit{``Sure, please enter the response from tutor''} & \textbf{Assistant} & \textit{``Please provide some examples of correct and incorrect response''} \\
\textbf{User}      & \texttt{\{Textual response\}}              & \textbf{User}      & \texttt{\{Correct example\}} + \texttt{\{Incorrect example\}}                            \\
                   &                                               & \textbf{Assistant} & \textit{``Sure, please enter the response from tutor''}                   \\
                   &                                               & \textbf{User}      & \texttt{\{Textual response\}}                                      \\ \hline
\end{tabular}
\end{table}

The prompt strategies are in the form of \textit{Chat-Completion}, which refers to the generated response produced by the GPT-4 model during a conversation. When a user provides a prompt, GPT-4 processes the prompt and generates a relevant response, known as the \textit{``Completion''}. The \textit{Chat-Completion} is set up to generate the label for each trainee's response. For Zero-shot implementation, as presented in Table \ref{tab:promt_bin}, the \textit{Chat-Completion} has three different chat roles: \textbf{System}, \textbf{User}, and \textbf{Assistant}. The role of \textbf{System} represents the assigned default character for the machine. In our case, GPT-4 facilitates the role of a \textit{``binary classifier''}. The role of \textbf{User} represents human input. The role of \textbf{Assistant} denotes a machine-generated response, which is to frame the prompting process as a conversation. Compared to the Zero-shot learning approach, the few-shot learning approach provides a limited number of correct and incorrect examples for the GPT-4 model to understand the classification patterns (Table \ref{tab:promt_bin}). Subsequently, our proposed prompt requires specific inputs from the \textbf{User}. The input of \texttt{\{Lesson Principle\}} is based on the principles of a correct response from the lesson materials created by \citeauthor{thomas2023tutor}~\cite{thomas2023tutor}. The input of \texttt{\{Textual response\}} is the trainee's response. As there are three distinct lessons, the input of \texttt{\{Lesson Name\}} in the instruction prompt is substituted with the appropriate lesson name.

\subsection{Enhancing the trainee responses by GPT models}

To explore \textbf{RQ2}, we used the GPT-4 model to rephrase incorrect responses into correct forms effectively. We designed the prompt strategies presented in Table \ref{tab:promt_rephrae}. For the Zero-shot learning, we assigned a role with GPT-4 to rephrase the trainee's response (i.e., ``\textit{You are rephrasing tutor’s response}''). For the role of \textbf{User}, similar to RQ1, we used \texttt{\{Lesson Principle\}} to enable GPT-4 to understand the correct form of tutor responses. To effectively rephrase the trainees' responses, we believe that providing context about the scenario in which the responses were given might lead GPT-4 to generate more accurate rephrased outputs. Thus, in the prompt, we also added the input of \texttt{\{Lesson Scenario\}}, which was the actual text of the scenario-based question, as demonstrated in Table \ref{tab:sample_praise}, \ref{tab:sample_error}, \& \ref{tab:sample_know}. In the context of the few-shot learning approach, we supplied two examples of rephrased incorrect responses in their correct forms provided in the training lessons to help the GPT-4 model infer the rephrasing rules (see Table \ref{tab:promt_bin}). The GPT-4 \textit{Chat-Completion} is presented in Table \ref{tab:promt_rephrae}.

\begin{table}[]
\caption{Prompt strategies for binary classifier. We used \textit{Chat-Completion} to process the trainees' responses in batch}
\footnotesize
\label{tab:promt_rephrae}
\renewcommand{\arraystretch}{1.5}
\begin{tabular}{|p{1.2cm}p{4.4cm}|p{1.2cm}p{4.4cm}|}
\hline
\multicolumn{2}{|l|}{\textbf{Zero Shot}}                                 & \multicolumn{2}{l|}{\textbf{Few-shot}}                                                       \\ \hline
\textbf{Role}      & \textbf{Content}                           & \textbf{Role}      & \textbf{Content}                                               \\ \hline
\textbf{System}    & \textit{``You are rephrasing tutor's response.''} & \textbf{System}    & \textit{``You are rephrasing tutor's response.''}                                  \\
\textbf{User} &
\texttt{\{Lesson Principle\}}+\textit{``The provided response attempts to answer to the following scenario.''}+\texttt{\{Lesson Scenario\}}+\textit{``Please rephrase the tutor's response according to the principle mentioned above to create a better example of''}+\texttt{\{Lesson Name\}}+\textit{``Retain words and ideas from the tutor’s response. Limit changes to the original tutor's response to a minimum. Maintain the same length as the original tutor's response. Please rephrase as less words as possible from the original tutor's response. Highest priority is to make sure to follow the principle of the correct response when rephrasing.''}&
  \textbf{User} & 
\texttt{\{Lesson Principle\}}+\textit{``The provided response attempts to answer to the following scenario.''}+\texttt{\{Lesson Scenario\}}+\textit{``Please rephrase the tutor's response according to the principle mentioned above to create a better example of''}+\texttt{\{Lesson Name\}} + \textit{``Retain words and ideas from the tutor’s response. Limit changes to the original tutor's response to a minimum. Maintain the same length as the original tutor's response. Please rephrase as less words as possible from the original tutor's response. Highest priority is to make sure to follow the principle of the correct response when rephrasing.''}
    \\
\textbf{Assistant} & \textit{``Sure, please enter the response''}            & \textbf{Assistant} & \textit{``Please provide some examples of how you will rephrase the given incorrect response to make it correct''} \\
\textbf{User}      & \texttt{\{Textual response\}}                          & \textbf{User}      & \texttt{\{Rephrased examples\}}                                             \\
                   &                                            & \textbf{Assistant} & \textit{``Sure, please enter the response''}                               \\
                   &                                            & \textbf{User}      & \texttt{\{Textual response\}}                                              \\ \hline
\end{tabular}
\end{table}




\subsection{Evaluation approach}
\label{eval_approach}

\textbf{Evaluation for RQ1.} We employ both the F1 score and the Area under the ROC curve (AUC) for evaluating the performance of our classification model. Furthermore, given our specific focus on identifying incorrect feedback, we incorporate two additional metrics: the Negative Predictive Value (NPV) and the True Negative Rate (TNR). These measures are crucial for determining the model's efficacy in minimizing false negatives and minimizing such errors is critical, as a false identification can result in incorrect feedback. Incorrect feedback can further undermine the training's effectiveness, potentially eroding trust and changing how trainees engage with the training activities. We provide the formulas for NPV and TNR in equations \ref{npv} and \ref{tnr}, respectively. Both NPV and TNR are metrics that range from 0 to 1, with higher values signifying a model's enhanced capability to correctly identify true negative instances.

\begin{equation}
\small
\label{npv}
\begin{aligned}
Negative \, Predictive \, Value \, (NPV) = \frac{True \, Negative}{True \, Negative + False \, Negative}
\end{aligned}
\end{equation}

\begin{equation}
\small
\label{tnr}
\begin{aligned}
True \, Negative \, Rate \, (TNR) = \frac{True \, Negative}{True \, Negative + False \, Positive}
\end{aligned}
\end{equation}

\vspace{1em}

\textbf{Evaluation for RQ2.} After rephrasing the trainee's responses, we evaluate the accuracy and quality of the rephrased responses. In order to achieve this, we first utilized the most effective binary classifier developed in RQ1 to classify the rephrase responses. Then, we compared the number of correct responses in rephrased responses and correct responses in original responses. Specifically, we wanted to investigate the extent to which the GPT-4 model has the capability to improve the accuracy of the trainee's responses. When the number of correct labels in rephrase responses is more than the correct responses in the original responses, it indicates that the GPT-4 model has the ability to accurately rephrase the trainee's responses and the classifier developed in RQ1 generally satisfied with the rephrased result. Additionally, we aim to compare the quality rephrased responses by GPT-4 with the ones by human expert. To do so, we first hired three experienced human tutors who completed the training for the three lessons. These three experts were asked to rephrase the incorrect responses based on the research recommendation provided in the lessons. Afterward, we invited a fourth human educational expert to assess the quality of rephrased responses in two dimensions: \textit{Accuracy} and \textit{Responsiveness}. The dimension of \textit{Accuracy} was used to measure the correctness of the rephrased responses. Regarding the dimension of \textit{Responsiveness}, it evaluates how the rephrased response selectively changes some words to improve the trainee’s original response, while largely preserving the original words and ideas from the trainee's response. In our study, we designed the question for evaluating \textit{Accuracy} by asking ``\textit{The rephrased response is a better example of \texttt{\{Lesson Name\}} than the trainee’s response}'' and the question for evaluating \textit{Responsiveness} by asking ``\textit{The rephrased response changes some words to improve the trainee's response, but otherwise keeps words and ideas from the trainee's response}''. The educational expert answered the questions by using the five-point Likert scale (i.e., \textit{Strongly Disagree} to \textit{Strongly Agree}).


\section{Results}
\label{result}

\subsection{Results for RQ1: Binary Classifier for Correct Responses }\label{rq1}

For RQ1, we explored the zero-shot and few-shot approaches to train a binary classifier using the GPT-4 model, as detailed in Sec.~\ref{method:rq1}. The classifier's performance is presented in Table \ref{tab:rq1_result}. For the lesson \textit{Giving Effective Praise}, the zero-shot approach resulted in an F1 score of 0.761 and an AUC of 0.743. When leveraging a two-way few-shot learning approach, we observed an improvement in the performance. The F1 scores remained consistently high, ranging from 0.856 to 0.872, with the 3-shot model achieving the peak performance. In parallel, the AUC scores were also robust, varying from 0.851 to 0.865, with the 5-shot model outperforming the others. Despite these improvements, the NPV and TNR metrics showed greater variability. The NPV spanned from 0.8 to 0.88, with the 3-shot model again taking the lead, whereas the TNR fluctuated between 0.744 to 0.851, with the 5-shot configuration achieving the strongest performance.

\begin{table}[!ht]
\caption{Classification performance of the responses from three lessons}
\label{tab:rq1_result}
\footnotesize
\renewcommand{\arraystretch}{1.3}
\begin{tabular}{@{}llcccccc@{}}
\toprule
\multirow{2}{*}{\textbf{Lessons}} &
  \multirow{2}{*}{\textbf{Metrics}} &
  \multirow{2}{*}{\textbf{Zero Shot}} &
  \multicolumn{5}{c}{\textbf{Two-way Few-shot}} \\ \cmidrule(l){4-8} 
 &
   &
   &
  \textbf{1-shot} &
  \textbf{2-shot} &
  \textbf{3-shot} &
  \textbf{4-shot} &
  \textbf{5-shot} \\ \midrule
\multirow{4}{*}{\textit{Giving Effective Praise}} &
  \textbf{F1} &
  0.761 &
  0.870 &
  0.845 &
  \textbf{0.872} &
  0.856 &
  0.860 \\
 &
  \textbf{AUC} &
  0.743 &
  0.858 &
  0.836 &
  0.863 &
  0.851 &
  \textbf{0.865} \\
 &
  \textbf{NPV} &
  0.666 &
  0.841 &
  0.853 &
  \textbf{0.881} &
  0.841 &
  0.800 \\
 &
  \textbf{TNR} &
  0.680 &
  0.787 &
  0.744 &
  0.787 &
  0.787 &
  \textbf{0.851} \\ \midrule
\multirow{4}{*}{\textit{Reacting to Errors}} &
  \textbf{F1} &
  0.767 &
  0.779 &
  0.821 &
  0.840 &
  0.823 &
  \textbf{0.867} \\
 &
  \textbf{AUC} &
  0.768 &
  0.778 &
  0.819 &
  0.838 &
  0.822 &
  \textbf{0.866} \\
 &
  \textbf{NPV} &
  \textbf{0.911} &
  0.892 &
  0.866 &
  0.857 &
  0.823 &
  0.880 \\
 &
  \textbf{TNR} &
  0.585 &
  0.622 &
  0.736 &
  0.792 &
  0.792 &
  \textbf{0.830} \\ \midrule
\multirow{4}{*}{\textit{\begin{tabular}[c]{@{}l@{}}Determining What \\ Students Know\end{tabular}}} &
  \textbf{F1} &
  0.660 &
  0.712 &
  0.718 &
  0.747 &
  0.798 &
  \textbf{0.805} \\
 &
  \textbf{AUC} &
  0.668 &
  0.712 &
  0.719 &
  0.748 &
  0.799 &
  \textbf{0.806} \\
 &
  \textbf{NPV} &
  0.630 &
  0.714 &
  0.733 &
  0.733 &
  0.818 &
  \textbf{0.821} \\
 &
  \textbf{TNR} &
  \textbf{0.828} &
  0.714 &
  0.786 &
  0.785 &
  0.771 &
  0.786 \\ \bottomrule
\end{tabular}
\footnotesize Note: \textbf{AUC} represents Area under the ROC Curve; \textbf{NPV} represent Negative Predicted Value; \textbf{TNR} represents True Negative Rate.
\end{table}

For the lesson on \textit{Reacting to Errors}, the performance of the zero-shot learning approach resulted in an F1 score of 0.767 and an AUC of 0.768. It is worth noting that the zero-shot learning approach had an impressive NPV score of 0.911, the highest NPV score for feedback from \textit{Reacting to Errors} activity, indicating the model's robustness in identifying true negative outcomes. When utilizing two-way few-shot learning approaches, the 5-shot learning approach presented the highest F1, AUC, and TNR scores at 0.867, 0.866, and 0.83, respectively. 

Lastly, for the lesson on \textit{Determining What Students Know}, the zero-shot learning approach resulted in an F1 score of 0.66 and AUC of 0.668, the lowest across the three lessons. Interestingly, the zero-shot model had a higher TNR score of 0.828, indicating that the model was adept at identifying true negative cases for this lesson. The performance across the F1, AUC, and NPV metrics presented a general uptick with the adoption of the two-way few-shot learning method, with the 5-shot variant demonstrating the highest enhancements, reflected by F1, AUC, and NPV scores of 0.805, 0.806, and 0.821, respectively.


\subsection{Results for RQ2: Using GPT-4 to Rephrase Incorrect Responses }\label{result_rq2}

For RQ2, we examine the application of GPT-4 in transforming trainees' incorrect responses into a preferred format that exemplifies effective feedback, thereby demonstrating the correct manner to meet learner needs through feedback revision. To accomplish this, we utilized the most effective binary classifier identified from RQ1, the 5-shot classifier, to pinpoint incorrect responses within the three lessons. The identified responses were then compared with the responses identified by the expert human raters as described in Sec.~\ref{method:rq1}. The intersection of the responses identified as incorrect by both the classifier and the human rates resulted in 36 responses for \textit{Giving Effective Praise}, 42 responses for \textit{Reacting to Errors}, and 53 responses for \textit{Determining What Students Know}. The overlap between the five-shot classifier and human raters was 85\%, 83\%, and 78.6\% for \textit{Giving Effective Praise}, \textit{Reacting to Errors}, and \textit{Determining What Students Know}, respectively, as indicated by the TNR scores for the 5-shot approach shown in Table~\ref{tab:rq1_result}.

As each training activity across the three lessons contained two paired examples to illustrate effective feedback in each scenario, we utilized the two paired examples per lesson to take a two-shot learning approach in exploring the effectiveness of GPT-4 in rephrasing student feedback. In this section, we report on the accuracy and responsiveness of the rephrased trainee responses by comparing the responses generated using zero-shot and two-shot GPT-4 models with responses rephrased by humans across the three lessons. The responses were assessed using a five-point Likert scale, i.e., \textit{Strongly Disagree} (represented by -2), \textit{Disagree} (represented by -1), \textit{Neutral} (represented by 0), \textit{Agree} (represented by 1), and \textit{Strongly Agree} (represented by 2), as described in Sec. \ref{eval_approach}. Given the ordinal nature of Likert scale data, we utilize the Mann-Whitney U test, a non-parametric statistical method, to ascertain if the accuracy and responsiveness of the rephrased responses are statistically different.

\begin{figure*}[b]
\centering
  \includegraphics[width=1\textwidth]{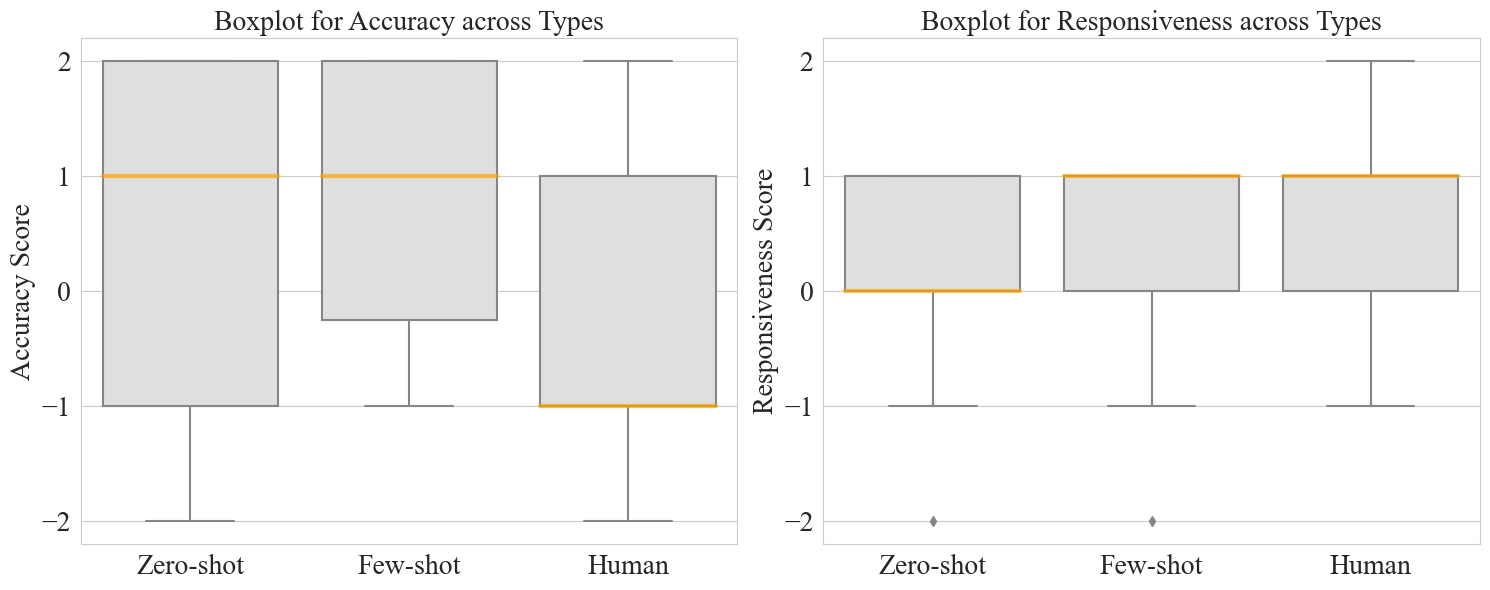}
\caption{Distribution of accuracy and responsiveness scores from the lesson \textit{Giving Effective Praise}}
\label{fig:praise_result}
\vspace{-6mm}
\end{figure*}

\begin{table}[b]
\caption{Statistics for rephrased responses from the lesson \textit{Giving Effective Praise}.}
\footnotesize
\renewcommand{\arraystretch}{1.3}
\label{tab:praise_stat}
\begin{tabular}{lccc}
\hline
\textbf{Metrics}       & \textbf{Zero-shot} & \textbf{Few-shot} & \textbf{Human} \\ \hline
Mean Accuracy          & 0.61               & 0.92              & -0.36           \\
Mean Responsiveness    & 0.22               & 0.44              & 0.44           \\
words/response (mean) & 17.28              & 21.72             & 12.28          \\
words/response (SD) & 6.28               & 18.94             & 5.88           \\ \hline
\end{tabular}
\end{table}

First, we examined the accuracy and responsiveness of the rephrased trainee responses for the lesson Giving Effective Praise, as presented in Fig. \ref{fig:praise_result}. We observed a higher median accuracy score of 1 for responses rephrased by GPT-4 (both Zero-shot and Few-shot) whereas the human rephrased responses received a median score of -1. As shown in Table \ref{tab:praise_stat}, the accuracy scores of the rephrased responses generated using both GPT models (zero-shot and few-shot) were significantly higher than the responses rephrased by the humans ($p <0.001$) indicating that the GPT-4 models were more effective at rephrasing the responses to the desired format in comparison to humans. While we did not observe a significant difference in the accuracy of the two GPT-based models, we observed a higher variance in the score of the zero-shot approach in comparison to the accuracy scores for the two-shot approach. When analyzing the responsiveness of the rephrased responses, we did not observe a significant difference between the responsiveness score of the GPT-4 rephrased responses and human rephrased responses; however, the human rephrased responses had a higher variance in comparison to the responsiveness scores of GPT-4 rephrased responses. The result demonstrated that the few-shot learning approach performed significantly better than the human in terms of the accuracy of the rephrased responses, while there was no significant difference in the responsiveness of the rephrased responses between the rephrased responses from the humans and the GPT-4 models. It indicated the effectiveness of few-shot learning on rephrasing the incorrect trainees' responses on the lesson of \textit{Giving Effective Praise}. 

\begin{figure*}[h]
\centering
  \includegraphics[width=1\textwidth]{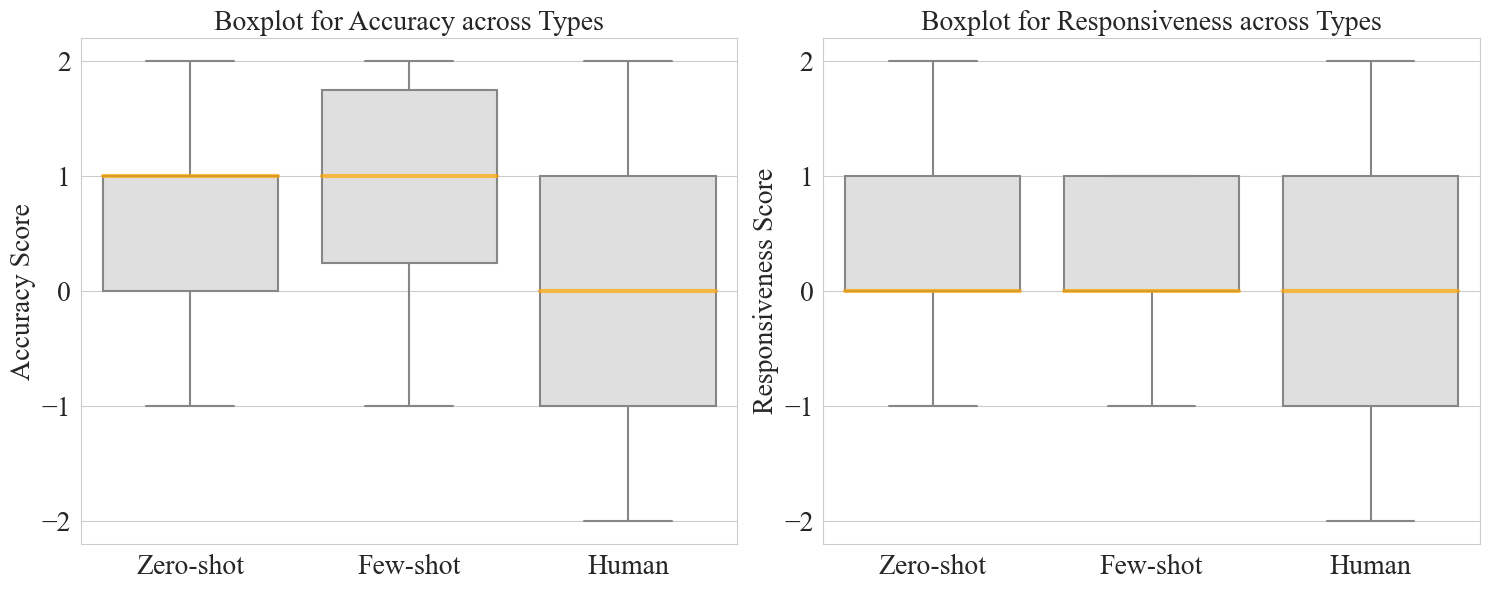}
\caption{Distribution of accuracy and responsiveness scores from the lesson \textit{Reacting to Errors}}
\vspace{-6mm}
\label{fig:error_result}
\end{figure*}

Similarly, we evaluated the rephrased responses provided by both GPT-4 models and human for the \textit{Reacting to Errors} lesson, presented in Fig. \ref{fig:error_result}. The GPT-4-generated responses achieved a median accuracy score of 1, outperforming the human-revised responses, which held a median score of 0. Upon examining the rating further, as presented in Table \ref{tab:error_stat}, the accuracy of responses rephrased using the few-shot approach was significantly higher than those rephrased by humans ($p<0.01$). Even the zero-shot rephrased responses were more accurate than human alterations ($p<0.05$). As for the responsiveness, most of the scores from the GPT-revised and human-revised responses were clustered between 0 and 1, with no significant difference in responsiveness between them. Additionally, the table also indicated that the average word count per response remained consistent between the GPT and human revisions, demonstrating that the GPT models, especially the few-shot approach, are adept at effectively rephrasing incorrect responses to \textit{Reacting to Errors} without extensive modification to the original wording and sentence structure provided by the trainees.

\begin{table}[h]
\caption{Statistics for rephrased responses from the lesson \textit{Reacting to Errors}.}
\footnotesize
\renewcommand{\arraystretch}{1.3}
\label{tab:error_stat}
\begin{tabular}{lccc}
\hline
\textbf{Metrics}       & \textbf{Zero-shot} & \textbf{Few-shot} & \textbf{Human} \\ \hline
Mean Accuracy          & 0.62               & 0.86              & 0.07           \\
Mean Responsiveness    & 0.17               & 0.17              & 0.21           \\

 words/response (mean) & 15.79              & 15.40             & 15.26          \\
words/response (SD) & 6.93               & 7.24              & 5.35           \\ \hline
\end{tabular}
\vspace{-3mm}
\end{table}

\begin{figure*}[!ht]
\centering
  \includegraphics[width=1\textwidth]{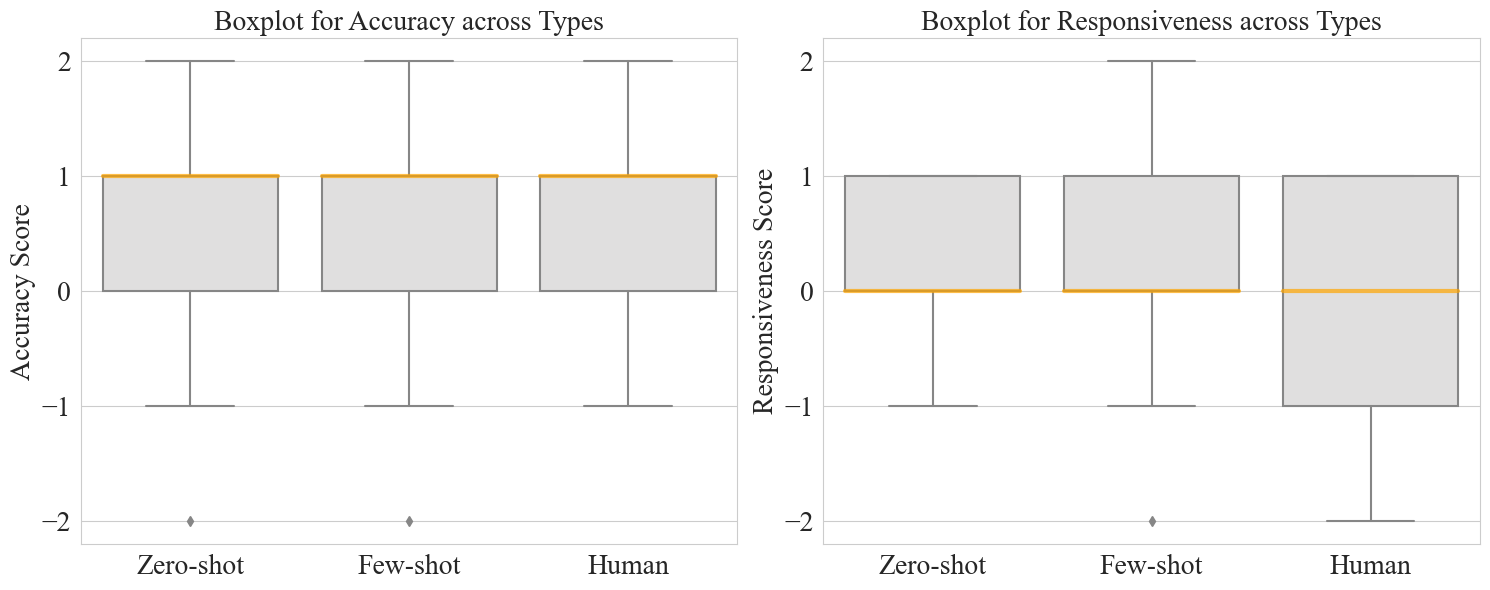}
\caption{Distribution of accuracy and responsiveness scores from the lesson \textit{Determining What Students Know}}
\vspace{-12mm}
\label{fig:know_result}
\end{figure*}

Finally, our evaluation of the rephrased responses from the lesson \textit{Determining What Students Know}, as illustrated in Fig. \ref{fig:know_result} and Table \ref{tab:know_stat}, revealed no significant difference in the dimensions of accuracy and responsiveness across the three approaches. Notably, unlike the accuracy in the other two chapters, the responsiveness scores from the few-shot method were marginally higher than those rephrased by humans ($p = 0.08$), indicating comparable performance between the automated few-shot and zero-shot approaches and human expertise. At the same time, no statistical significance was observed across conditions for responsiveness. Interestingly, it was in the \textit{Determining What Students Know} lesson that the classification model had its weakest performance among the three lessons. 


\begin{table}[!ht]
\caption{Statistics for rephrased responses from the lesson \textit{Determining What Students Know}.}
\footnotesize
\renewcommand{\arraystretch}{1.3}
\label{tab:know_stat}
\begin{tabular}{lccc}
\hline
\textbf{Metrics}       & \textbf{Zero-shot} & \textbf{Few-shot} & \textbf{Human} \\ \hline
Mean Accuracy          & 0.68               & 0.70              & 0.85           \\
Mean Responsiveness    & 0.28               & 0.30              & 0.06           \\
words/response (mean) & 22.72              & 20.83             & 20.09          \\
words/response (SD) & 18.51              & 16.12             & 8.10           \\ \hline
\end{tabular}
\vspace{-3mm}
\end{table}

\section{Discussion}\label{sec12}

Providing explanatory feedback is a fundamental requirement for delivering personalized feedback to learners. Our study explored the use of large language models (GPT-4 model) to automate the facilitation of explanatory feedback to novice tutors, where the main findings can be summarized in two folds:

\textit{Firstly}. GPT-4 models, especially for the few-shot approach, have the potential to accurately identify the correct and incorrect trainees' responses, which can be used to provide corrective feedback when training novice tutors on the scenario-based tasks. Our results indicate that despite a limited number of samples, the GPT-4 model can accurately identify the incorrect trainees' responses across three different tutor training lessons (i.e., \textit{Giving Effective Praise}, \textit{Reacting to Errors}, and \textit{Determining What Students Know}). By comparing the classification performance with zero-shot learning, the few-shot learning approach, especially with increasing shots, generally tends to improve the model's classification performance. This improvement suggests that more examples might increase GPT's capability to recognize the many different ways to express a target concept like effort-based praise (e.g., \textit{``Good effort on solving the problem''}), and distinguish it from a related concept, like outcome-based praise (e.g., \textit{``Good job''}). The implications of this finding is profound, especially when considered alongside existing research on neural network learning in humans. Previous research \cite{carvalho2022computational} has illustrated that both the quantity and diversity of examples play a significant role in the learning process, with optimal outcomes achieved through exposure to a range of examples that are internally diverse yet distinct from other categories. Applying this principle to the context of LLM training suggests a strategy where examples within a category (e.g., praising effort) are maximally diverse, whereas examples across categories are closely aligned (e.g., comparing praise for effort with praise for outcomes). Pursuing this line of inquiry in future research could yield valuable insights into the mechanisms underpinning effective learning in both human and artificial neural networks. By systematically exploring the interplay between example diversity and learning efficacy, we can refine our understanding of how best to structure training data for LLMs like GPT-4, ultimately enhancing their utility in educational applications.




\textit{Secondly}, the capability of GPT-4, particularly when employing the few-shot learning approach, extends to effectively rephrasing trainees' incorrect responses into a desired format. Notably, GPT-4's performance in rephrasing incorrect responses to correct ones is on par with, and sometimes surpasses, that of experienced human tutors. This proficiency likely stems from GPT-4's advanced understanding of context and language nuances \cite{openai2023gpt4}, enabling it to reconstruct trainees' incorrect responses to align more closely with the desired responses. The practical implications of the GPT4's capabilities are significant. The classified and rephrased responses generated by GPT-4 can be integrated into template-based feedback systems. Such integration facilitates the provision of real-time and explanatory feedback to novice tutors (or trainees) during their training sessions.




\subsection{Implications}

The incorporation of the binary classifier and its generalizability in terms of performance holds significant implications for providing explanatory feedback. The classified results (i.e., correct or incorrect responses) on trainee's responses can be further integrated into the provision of corrective feedback as shown in Fig. \ref{fig:intro_tem_feedback}. Specifically, by identifying the incorrect responses, our feedback system can use the template-based feedback to provide suggestions for trainees to consider, i.e., ``\textit{AI-generated feedback suggests that your response could focus more on praising the student for their efforts in the learning process.} Providing corrective feedback is essential in the learning process for tutor training since it can assist the tutors in identifying their errors and improving the quality of their feedback~\cite{butler2013explanation}.

Furthermore, this study demonstrated the potential of prompting GPT-4 models in rephrasing incorrect trainees' responses into the desired form. We measured the quality of rephrased responses from GPT-4 models and human experts in terms of their accuracy and responsiveness as described in Sec. \ref{eval_approach}. Based on our observations, the rephrased responses consistently rated higher in accuracy while the responsiveness of the GPT-4 generated responses were comparable to the human-generated responses. For instance, a tutor response classified as incorrect, \textit{``\underline{Great Job}! But there is a tiny \underline{mistake}, can you tell what was it ?''} from the lesson \textit{Reacting to Errors} was rephrased in the following manner:
\vspace{1mm}
\begin{itemize}
    \item[] \textbf{Zero-shot:} \textit{``\underline{Good effort}, Aaron! There's a \underline{small detail that needs tweaking.} Can you spot what it might be?''}
    \item[] \textbf{Few-shot:} \textit{``\underline{Good effort!} \underline{Let's take another look at this}, can you walk me through your process?''} 
    \item[] \textbf{Human:} \textit{``\underline{Great Job}! But there is \underline{a tiny mistake}, can you tell what was it by going over each step?''}
\end{itemize}
\vspace{1mm}

The incorrect response used the word ``\textit{\underline{mistake}}'' in the response, which failed to reframe the mistakes as a learning opportunity effectively and is considered to be an ineffective tutoring strategy as indicated by \cite{chhabra2022evaluation, thomas2023tutor}. In contrast, both zero-shot and few-shot responses were more effective in accounting for the student's errors when rephrasing the feedback. The zero-shot rephrased response avoids using direct words about the student’s mistake and implicitly mentions the misconception ``\textit{There's a small detail that needs tweaking.}'' whereas the few-shot approach encourages the student to revisit the problem \textit{\underline{Let's take another look at this}} which is also in line with the effective practice of reacting to errors~\cite{chhabra2022evaluation, thomas2023tutor}. It should be noted that both zero-shot and few-shot rephrased responses also changed the praise on student's outcome (i.e., ``\textit{\underline{Great Job}}'') to student learning effort (i.e., ``\textit{\underline{Good effort}}''). However, the human expert failed to rephrase the incorrect response into the desired form even though the expert changed the words from the original incorrect response. Additionally, when the accuracy of rephrased responses was rated at 1 (i.e., \textit{Agree} on a five-point Likert scale) or 2 (i.e., \textit{Strongly Agree}) but the responsiveness was rated from -2 (i.e., \textit{Strongly Disagree}) to 0 (i.e., \textit{Neutral}), the rephrased responses still were presented effective responses. For example, an example of incorrect response \textit{``I think you are \underline{doing great.}''} from \textit{Giving Effective Praise} was rephrased as below:

\vspace{1mm}
\begin{itemize}
    \item[] \textbf{Zero-shot:} \textit{``You're showing \underline{excellent effort} in tackling your math work, Carla. Your \underline{perseverance is commendable}.''} 
    \item[] \textbf{Few-shot:} \textit{``I see you're making \underline{good progress with your efforts}. Keep it up!''} 
    \item[] \textbf{Human:} \textit{``You are doing a \underline{great job!}''} 
\end{itemize}
\vspace{1mm}

\begin{figure*}[!ht]
\centering
  \includegraphics[width=0.48\textwidth]{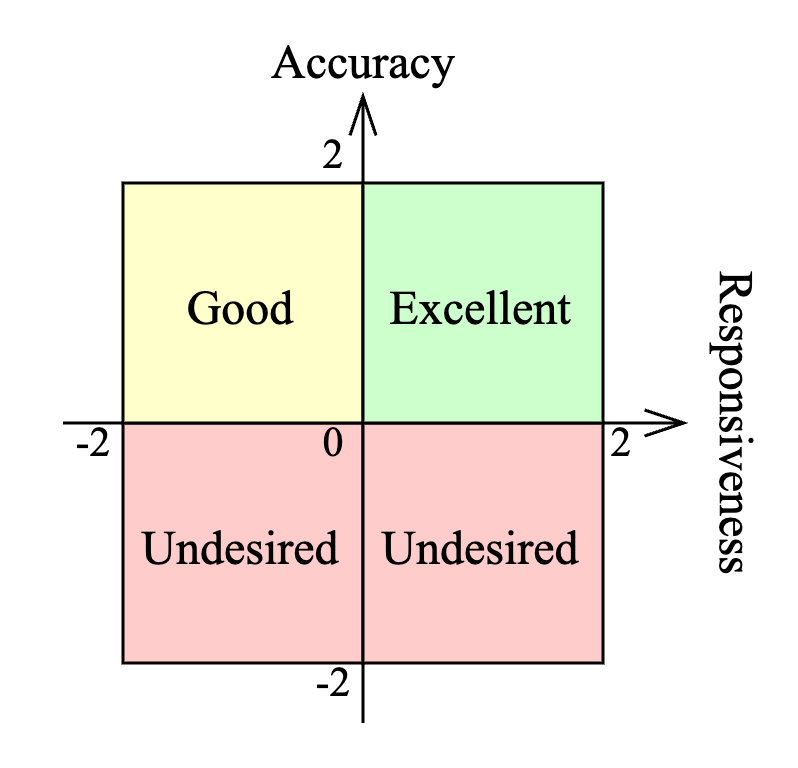}
\caption{Framework for determining the quality of the rephrased responses.}
\label{fig:framework}
\vspace{-3mm}
\end{figure*}

The incorrect response \textit{``I think you are \underline{doing great}.''} failed to praise student on their learning efforts or learning actions but on their learning outcome, which is considered ineffective praise as indicated in~\cite{chhabra2022evaluation, thomas2023tutor}. Both zero-shot and few-shot rephrased responses were rated accuracy of 2 (i.e., \textit{Strongly Agree}) but responsiveness of 0 (i.e., \textit{Neutral}). Both shot and few-shot rephrased responses demonstrated praise on the student learning efforts as underlined in the examples, but both responses changed many words, which was not similar to the original incorrect responses. In comparison, the responsiveness of human rephrased responses was rated at 1 since there was only several words were changed from the original incorrect response. However, the human expert failed to revise the praise correctly, and the rephrased response was rated at -1 (i.e., \textit{Disagree}). The rephrased praise still focused on the student learning outcome (i.e., \textit{\underline{great job!}}) rather than their learning efforts, which is not considered an effective response for praising student as indicated by~\cite{thomas2023tutor}. As summarized by the evaluation results of both GPT-4 and human rephrased responses, we proposed a framework for determining the quality of the rephrased responses, shown in Fig. \ref{fig:framework}. 

This framework (Fig. \ref{fig:framework}) aims to guide future work to understand the extent to which the rephrased responses are considered high quality. When the accuracy of the rephrased response is rated at 1 or 2, the rephrased responses are considered to be acceptable. Based on our observation, the optimal rephrased responses should be high in both accuracy and responsiveness (i.e., \textbf{Excellent} area in Fig. \ref{fig:framework}), which could guide the trainees to understand the desired form of the responses and also help them know where they did not perform well while providing their scenario specific feedback. Since the dimension of responsiveness aims to minimize the changes of words in the responses, we expect the trainees to be able to locate the parts of the sentence that are incorrect and rephrase them accordingly. Similarly, a high accuracy and lower responsiveness (i.e., \textbf{Good} area in Fig. \ref{fig:framework}) could guide the trainee to recognize the desired quality of the feedback. However, as shown in the above example, the low responsiveness of the rephrased responses is an indicator of the modifications required in the original incorrect responses, which may not be as helpful to the trainees if the rephrasing resulted in major structural and semantic changes that are harder to learn and retain. Finally, we defined responses in two areas as undesirable responses, as illustrated in Fig. \ref{fig:framework}. The undesirable responses, marked by a low accuracy score ($\leq 0$), undermine the effectiveness of the feedback~\cite{thomas2023tutor}. While the rephrased responses might demonstrate high responsiveness, the low accuracy of the response is still detrimental to its effectiveness and, as such, is not desirable. The rephrased feedback (\textit{``You are doing a \underline{great job!}''} ), as presented above, is an example of a rephrased response with a low accuracy but high responsiveness score.

\subsection{Limitations and Future Work}
\label{future}


\noindent \textbf{Evaluating impact of proposed feedback system on tutoring practice.} While our current findings demonstrated the potential of GPT models in providing explanatory feedback and appropriately rephrased responses, there is a need for a more comprehensive evaluation of such feedback's effectiveness in tutor training. In future work, we plan to investigate the influence of the feedback on tutor practice. Specifically, we will examine the direct effects of our feedback on tutors' skill acquisition, retention, and application in real-world tutoring scenarios. By conducting longitudinal studies with both control and experimental groups, we aim to gain a clearer understanding of the long-term advantages and possible challenges of our approach. Such insights will not only shed light on the efficacy of our feedback system but also inform potential refinements to enhance the training process for novice tutors.

\vspace{1.5mm}

\noindent \textbf{Using advanced prompt strategies for explanatory feedback.} In our current study, we utilized zero-shot and few-shot prompt strategies to identify correct or incorrect trainees' responses (RQ1) and to rephrase these incorrect responses appropriately (RQ2). While our proposed prompting strategies demonstrated promising results, there is potential for further improvement. We are considering the adoption of more advanced prompt strategies. Two such strategies that have caught our attention are the \textit{Tree of Thoughts}~\cite{yao2023tree} and \textit{Graph of Thoughts}~\cite{besta2023graph}. These prompting strategies are expected to offer a more nuanced and structured way of understanding the task context and generating relevant information, potentially leading to more accurate and insightful results. A comprehensive exploration of these advanced prompting strategies is beyond the scope of our current study. Thus, in future work, we aim to delve deeper into these prompt strategies to investigate their efficacy and potentials on the improvement of the quality of explanatory feedback.



\vspace{1.5mm}

\noindent \textbf{Generalizability across other tutor training lessons.} 
While our study demonstrated promising results on providing explanatory feedback primarily from three lessons, to further explore the efficacy of our feedback system, broader evaluations of the feedback system on other lessons are also important such as \textit{Using Motivational Strategies} and \textit{Ensuring Conceptual Understanding}. All the lessons on our platform introduces tutors to unique teaching scenarios and challenges. Ensuring that our feedback system is equally adept at handling the intricacies of each lesson is crucial for its overall success. Thus, it is important to evaluate the efficacy of our developed feedback system across all lessons, ensuring that the feedback provided is accurate, relevant, and conducive to the emerging tutor training process, continuously guiding tutors towards pedagogical excellence.

\vspace{1.5mm}

\noindent \textbf{Enhancing explanatory feedback through sequence labeling}
The primary objective of this study is to provide automatic explanatory feedback. We have demosntrated the demo of our developed explanaotry feedback system shown in Fig. \ref{fig:intro_tem_feedback}. To further unlock the potential of automatic explanatory feedback, we propose a significant enhancement: the integration of sequence labeling method, as originally introduced in the work by \cite{lin2023using, lin2024i}. In their research, they employed a color-coded highlighting approach to distinguish between the effective and ineffective component of trainee's responses, aiming to facilitate a clearer comprehension of correctness or incorrectness. By incorporating this sequence labeling approach in the provision of explanatory feedback, we expect that the feedback can demonstrate more corrective information fostering a deeper understanding among trainees regarding the construction of effective responses. 

\vspace{1.5mm}

\noindent\textbf{Enhancing trainee response evaluation beyond binary classification.} Our study leveraged GPT-4's capabilities to categorize trainee responses into binary classes: correct or incorrect. However, this dichotomous approach may be overly simplistic and potentially limiting for real-world applications where a more nuanced understanding is required. Acknowledging this, we recognize the necessity of developing a more granular evaluation scale. A tiered ranking system, perhaps on a five- or ten-point scale, could provide a more detailed and effective assessment of trainee responses, aligning more closely with the complexities of real-world scenarios. This insight highlights a limitation in our current methodology and underscores the potential for future research to explore more sophisticated classification frameworks that can capture the varied spectrum of trainee performance more accurately.

\vspace{1.5mm}

\noindent\textbf{Strategies for safeguarding privacy information in real-world tutoring.} Our study observed that responses from trainee tutors across three different lessons often included the use of student names, as in ``\textit{\underline{Kevin}, good job getting the problem correct!}'' This pattern suggests a tendency among some tutors to personalize their feedback by mentioning students by name during actual tutoring sessions. To further evaluate the practices of novice tutors within real-world tutoring contexts, it is necessary to colect and archive transcripts of tutoring dialogues in our database. To protect data privacy, we intend to anonymize any sensitive information, such as names, locations, and ages, contained within these transcripts.

\vspace{1.5mm}

\noindent\textbf{Enhancing automated explanatory feedback quality through human-in-the-loop design.} In our future work, we aim to explore the enhancement of automated explanatory feedback quality through the incorporation of a human-in-the-loop design. This approach will involve integrating human interaction directly into the feedback loop, enabling a ranking system where responses generated by Large Language Models (LLMs) are reviewed and prioritized based on human judgment. Such a mechanism is expected to provide stronger signals to the AI, guiding it towards producing outputs that are more aligned with human expectations.

\vspace{1.5mm}

\noindent \textbf{Crowd sourcing the evaluation of rephrased responses from trainees.} Inviting educational experts to evaluate the quality of rephrased responses is often time-consuming and impractical, especially when dealing with a large volume of tutor responses. To address this issue, we suggest a crowd-sourcing approach for rating the rephrased responses. we plan to include the question (shown in Table \ref{tab:future_work}) into the lesson and invite tutor trainees to answer the question. Table \ref{tab:future_work} presents the Scenario question and a response from a previous trainee which was identified an incorrect response. We will employ the large language models to rephrase the incorrect trainee's response and also keep the original incorrect response in the question. The new trainees are invited to rate the quality of responses based on the accuracy, responsiveness in a five-point scale. We also incorporate the original response for trainee to rate their scores. Since our developed binary classifier was not perfect, misclassified incorrect might exist, we also want the trainee's to provide their rating on the original responses. By doing so, we can obtain their ratings of rephrased responses and we expect our trainees can obtain better understanding about the presence of the effective form of responses in different training lessons.

\begin{table}[h]
\caption{Sample question for crowd sourcing the ratings of rephrased responses from the trainee tutors.}
\footnotesize
\renewcommand{\arraystretch}{1.5}
\label{tab:future_work}
\begin{tabular}{p{8cm}cc}
\hline
\multicolumn{3}{l}{\begin{tabular}[c]{@{}l@{}} \textbf{Scenario:} \textit{What exactly would you say to Cindy to begin helping her solve the math problem?} \\ \textbf{Response:} \textit{Are you familiar with the definition and notation for angle and side congruence?}\end{tabular}} \rule{0pt}{3.6ex}\rule[-2.2ex]{0pt}{0pt} \\ \hline
\textbf{Rephrased Responses}                                                                              & \textbf{Accuracy} & \textbf{Responsiveness}  \rule{0pt}{3.6ex}\rule[-2.2ex]{0pt}{0pt} \\ \hline
\textit{1. How would you define when angles or sides in a triangle are congruent? What does that mean to you?}        &    4      &       3                     \\ \hline
\textit{2. Can you explain your understanding of angle and side congruence, and their notations?}   &      4    &           -                      \\ \hline
\textit{3. What do you understand about the concept of congruence in relation to sides and angles?} &     -     &         -              \\  \hline
\textit{4. Are you familiar with the definition and notation for angle and side congruence?} &     -     &           -           \\  \hline
\end{tabular}
\end{table}

\vspace{1.5mm}

\noindent \textbf{Explanatory feedback for the synchronous tutoring session.} Our study demonstrated the capability of GPT-4 models to provide explanatory feedback and adeptly rephrasing tutor responses into a desired format. As shown in Sec. \ref{result_rq2}, our proposed few-shot learning approach could achieve performance comparable to human experts in rephrasing responses appropriately, which could help reduce the use of inappropriate instructional responses during the student learning process. Given our current findings, we expect the integration of our developed explanatory feedback system into synchronous text-based online tutoring could facilitate the tutoring process. Previous studies~\cite{lin2022good, lin2022exploring, lin2023role} have emphasized the importance of showing effective responses to students. Given the growing demand for qualified tutors, our feedback system, when integrated with synchronous tutoring platforms, can equip novice tutors to deliver timely and appropriate instructional feedback. To assess the influence of our exploratory feedback system on tutoring, We recommend conducting randomized controlled experiments to examine the efficacy of our feedback system further. In the experiment setup, tutors in experimental group will use our explanatory feedback system to provide instructional response, whereas the tutors in the control group will follow business-as-usual tutoring. The investigation aims for a comprehensive understanding of  the system's strengths and areas needing improvement.

\section{Conclusion}\label{conclusion}

We aimed to provide automatic explanatory feedback to enhance tutor training. Our study explored the potential of GPT-4 model in delivering real-time explanatory feedback for open-ended questions selected from three tutor training lessons. We first prompted the GPT-4 model to act as a binary classifier to identify incorrect tutor responses. With well-designed prompting strategies, the GPT-4 model, using a few-shot approach, accurately identified incorrect trainee responses across all three lessons we examined. We then used the GPT-4 model to rephrase incorrect responses into the desired responses. Our results demonstrated that the quality of rephrased responses provided by GPT-4, using a few-shot approach, achieved performance comparable to that of human experts. These results indicate that our proposed automatic explanatory feedback system shows promise in providing real-time feedback. Our study sheds light on the development of feedback provision for learners. By integrating our feedback system, we expect it can facilitate the tutor training process and further alleviate the the challenges associated with recruiting qualified tutors.


\backmatter


\bmhead{Acknowledgments} This work is supported by funding from the Richard King Mellon Foundation (Grant \#10851). Any opinions, findings, and conclusions expressed in this paper are those of the authors. We also wish to express our gratitude to Dr. Ralph Abboud for his invaluable guidance and recommendations, and to the members of Ken’s lab for their insightful feedback on this work. Special thanks to Ms. Jiarui Rao for her assistance in verifying the rating scheme.

\section*{Declarations}

\textbf{Ethics Approval} The study presented in this paper obtained the Institutional Review Boards (IRB) approval from Carnegie Mellon University.

\noindent\textbf{Conflicts of Interest} The authors have no relevant financial or non-financial interests to disclose, nor conflicting interests nor competing interests.

\bibliography{sn-bibliography}

\appendix

\end{document}